\documentclass[11pt]{article}

\usepackage[final]{acl}

\usepackage{times}
\usepackage{latexsym}

\usepackage[T1]{fontenc}

\usepackage[utf8]{inputenc}

\usepackage{microtype}

\usepackage{inconsolata}

\usepackage{makecell}
\usepackage{amsmath}

\usepackage{arydshln}
\usepackage{hyperref}
\usepackage[normalem]{ulem}
\usepackage{tabularray}
\usepackage{multirow}
\UseTblrLibrary{siunitx}

\usepackage{url}
\usepackage{todonotes}
\usepackage{enumitem}
\usepackage[capitalise]{cleveref}
\usepackage{rotating}
\usepackage{graphicx}

\usepackage{booktabs}
\usepackage{float}
\usepackage{placeins}  
\usepackage{lscape}  
\usepackage{wrapfig}

\usepackage{xcolor}
\usepackage{soul}
\sethlcolor{green} 

\title{CliniBench: A Clinical Outcome Prediction Benchmark for Generative and Encoder-Based Language Models}

\author{
 \textbf{Paul Grundmann\textsuperscript{1}},
 \textbf{Jan Frick\textsuperscript{1}},
 \textbf{Dennis Fast\textsuperscript{1}},
 \textbf{Thomas Steffek\textsuperscript{1}},
\\
 \textbf{Felix Gers\textsuperscript{1}},
 \textbf{Wolfgang Nejdl\textsuperscript{2}},
 \textbf{Alexander Löser\textsuperscript{1}},
\\
\\
 \textsuperscript{1}DATEXIS, Berlin University of Applied Sciences, Berlin, Germany
\\
 \textsuperscript{2}Leibniz University Hannover, Hannover, Germany
\\
 \small{
   \textbf{Correspondence:} \href{mailto:jan.frick@bht-berlin.de}{jan.frick@bht-berlin.de}
 }
}

\begin{document}
\maketitle
\begin{abstract}
With their growing capabilities, generative large language models (LLMs) are being increasingly investigated for complex medical tasks.
However, their effectiveness in real-world clinical applications remains underexplored.
To address this, we present CliniBench, the first benchmark that enables comparability of well-studied encoder-based classifiers and generative LLMs for discharge diagnosis prediction from admission notes in the MIMIC-IV dataset.
Our extensive study compares 12 generative LLMs and 3 encoder-based classifiers and demonstrates that encoder-based classifiers consistently outperform generative models in diagnosis prediction.
We assess several retrieval augmentation strategies for in-context learning from similar patients and find that they provide notable performance improvements for generative LLMs.
\end{abstract}

\section{Introduction}



The integration of transformer-based language models in clinical decision support systems is a rapidly advancing field, offering transformative potential for personalized healthcare, early diagnosis, and treatment planning \cite{laohawetwanit-medical-llm-survey-2024}. 
Among these applications, language models are being studied for the task of diagnosis prediction from admission notes, an extreme multi-label classification task with unbalanced data \cite{van-aken-etal-2021-clinical-outcome, van-aken-etal-2022-proto-patient, roehr-etal-2024-revisiting-core, grundmann-etal-2024-data-drift, figueroa-2024-sparse-proto-patient, gema_parameter-efficient_2024}.
Solving this machine learning task requires resources and expertise, which are often scarce or do not exist in hospitals. 
Furthermore, privacy and data protection laws often require that models are trained and applied within the hospital, preventing data or model sharing.

\paragraph{Encoder-based classifiers (Encoder)}
currently represent the state-of-the-art for diagnosis prediction, with most research focusing on fine-tuning pretrained models.
While they offer cost-efficient inference and strong task adaptation, their often fixed label space limits scalability, and they risk overfitting to training data.
Rapid developments in medical research and the emergence of new diseases necessitate the continual adaptation of existing models.
This is typically achieved via repeated fine-tuning, which in turn can lead to catastrophic forgetting.
These challenges have fueled a growing interest in generative large language models (LLMs).

\paragraph{Generative LLMs}
offer key benefits such as zero-shot inference, dynamic label spaces, and in-context learning (ICL) \cite{few_shot_gpt3}.
These features reduce the need for fine-tuning and result in a flexibility that encoder-based classifiers are missing.
Therefore, generative LLMs should be particularly well suited for clinical applications, where labeled datasets are costly to create due to data protection laws.
Despite their potential, generative models have not been systematically benchmarked against encoder-based classifiers for the diagnosis prediction task.

\paragraph{Learning at Test Time.}
While generative LLMs excel in adaptability and performance across diverse tasks, they are prone to hallucinations (i.e., factually incorrect or irrelevant outputs).
To mitigate this limitation, they can be augmented with techniques like Retrieval-Augmented Generation (RAG) \cite{lewis-2020-rag} and Chain-of-Thought (CoT) \cite{wei-2022-cot} prompting. 
RAG improves the reliability of generative models by linking them to external knowledge bases, improving the factual accuracy of their outputs. 
CoT prompting enhances logical consistency by encouraging step-by-step explanations, which helps with complex tasks.
CoT requires only minimal adjustments to the model's prompt structure, while RAG leverages existing hospital databases.
These strategies enable generative LLMs to become more robust and reliable, making them a practical choice for healthcare applications, even in resource-constrained environments.

\paragraph{Contributions.}
In this paper, we introduce Clini\-Bench, a comprehensive benchmark designed to simulate the unique challenges of clinical outcome prediction. CliniBench enables a systematic comparison of traditional state-of-the-art transformer encoders and emerging generative models.
We summarize our contributions as follows:

\begin{itemize}
    \item We publish the first benchmark\footnote{\url{https://github.com/datexis/clinibench}} to compare encoders with generative LLMs for diagnosis prediction from admission notes based on MIMIC-IV \cite{johnson_mimic-iv_2023}.
    \item We evaluate the augmentation techniques RAG and CoT prompting and compare them with the performance of state-of-the-art fine-tuned encoder models. 
    \item We provide an in-depth error analysis on a comprehensive set of experiments across various model configurations and augmentation strategies.
\end{itemize}

\section{Related Work}

\label{sec:related_work_encoder}

Clinical outcome prediction is based solely on information available at hospital admission represents a distinct challenge compared to retrospective ICD coding or predefined text classification tasks. 
\citet{van-aken-etal-2021-clinical-outcome} introduce the task based on MIMIC-III \cite{goldberger_physiobank_2000, johnson_mimic-iii_2016} to forecast discharge outcomes such as ICD codes, mortality, and length-of-stay. 
\citet{naik-2022-literature-augmented-outcome-prediction} propose enhancing outcome prediction by retrieving relevant biomedical literature, grounding the predictions to improve both accuracy and interpretability.
\citet{roehr-etal-2024-revisiting-core} extend this task to MIMIC-IV \cite{johnson_mimic-iv_2023}, adding new outcomes like patient routing and evaluating benchmark performance. \citet{grundmann-etal-2024-data-drift} examine data drift between MIMIC versions and highlight inconsistencies in ICD documentation practices, taxonomy, and label reliability.
To improve label generalization and interpretability, \citet{van-aken-etal-2022-proto-patient} propose ProtoPatient, a prototypical network-based model. This approach was later extended by \citet{figueroa-2024-sparse-proto-patient} to better handle long-tail ICD distributions. \citet{gema_parameter-efficient_2024} demonstrate that fine-tuning LLMs for classification, rather than generation, can yield improved performance on admission-based tasks using parameter-efficient methods.
\citet{shoham_cpllm_2024} focus on predicting readmission for three specific conditions, using LLMs trained on structured diagnosis descriptions from prior visits. In contrast to similar studies that utilize the full patient notes as inputs, the authors use textual descriptions of previously assigned diagnostic codes for their prediction.

\subsection{Related Tasks}
Two common well-studied, related tasks are ICD coding and medical question answering. Furthermore, as noted by \citet{bedi-2025-testing-and-evaluation-of-health-care-applications-of-llms}, only five percent currently make use of real patient records and realistic clinical tasks.

\paragraph{ICD coding}
resembles an information extraction rather than a diagnosis prediction task because it incorporates all available information about a patient at discharge time.
Related works include \citet{mullenbach-etal-2018-explainable}, \citet{nguyen-etal-2023-two-stage-decoder-icd-coding}, and \citet{nguyen2023mimicivicdnewbenchmarkextreme}, which focus on ICD coding of clinical notes from MIMIC-II, MIMIC-III and MIMIC-IV.
\citet{boyle_automated_2023} propose a tree-search method using the hierarchical structure of ICD codes, enabling superior performance on rare codes. Similarly, \citet{ong_applying_2023} evaluate ChatGPT for ICD coding in retina clinics, achieving 70\% accuracy but highlighting challenges such as hallucinations, underscoring the need for oversight in clinical applications.

\paragraph{Medical question answering.}
Beyond coding, LLMs are increasingly applied to medical question answering tasks. For instance, \citet{singhal-2024-towards-expert-level} and \citet{xie2024llamafoundationlargelanguage} evaluate closed-source models on curated diagnosis-ranking benchmarks using small numbers of cases. \citet{wu-etal-2024-knowlab} propose using chain-of-thought prompting and span correction for clinical note error detection.

\subsection{Learning at Inference Time}
To mitigate some of the general limitations of LLMs, researchers have explored various techniques such as chain-of-thought prompting \cite{wei-2022-cot}, few-shot learning \cite{parnami-2022-summary-few-shot-learning, song-2023-survey-few-shot-learning}, self-reflection \cite{asai-2024-selfrag, jeong-2024-medical-reasoning-rag-self-reflection}, self-verification \cite{weng-etal-2023-self-verification, gero_self-verification_2023} and guided decoding \cite{kuchnik-validating-llms-with-relm-2023, scholak-picard-2021, guided_generation-louf-2023}.
For our evaluation, we focus on RAG, CoT, and guided decoding to mitigate hallucinations.

\paragraph{Retrieval augmented generation}
enhances LLMs by incorporating external knowledge via in-context learning, improving accuracy, and reducing hallucinations.
For example, \citet{lu-etal-2024-clinicalrag} introduced ClinicalRAG for reliable clinical decision support, \citet{kresevic_optimization_2024} used RAG to increase guideline interpretation accuracy to 99\% and \citet{unlu-2024-rag-clinical-trial-screening} applied RAG for efficient clinical trial screening, outperforming manual methods. 
While \citet{xiong-2024-medical-rag-benchmarking} benchmarked RAG systems, showing up to 18\% accuracy improvements, \citet{ngo-2024-rag-evaluation} provided practical optimization guidelines for RAG systems. Finally, \citet{li-2025-biomed-rag} show that chunking and fine-granular data selection for RAG improve performance on biomedical relation extraction and classification tasks.

\paragraph{Chain-of-Thought prompting}
 enables step-by-step reasoning in LLMs, excelling in multi-step and symbolic tasks.
\citet{wei-2022-cot} introduce CoT and demonstrate its effectiveness in reasoning tasks. \citet{zhang-2023-automatic-cot} automate CoT prompt generation to scale its application.
\citet{wang-2023-cue-cot} tailor CoT for in-depth dialogue, enhancing coherence in responses.
\citet{sprague-2024-cot-analysis} analyze CoT's strengths in math and symbolic reasoning while highlighting its domain limitations.

\paragraph{Guided Decoding.}
To mitigate potential hallucinations that affect the parsability of model-generated output \citet{kuchnik-validating-llms-with-relm-2023, scholak-picard-2021} propose to limit the output token space of the LLM during generation given a finite state machine constructed from a regular expression and the respective context that has already been generated by the LLM.
Other approaches, such as prompting the model to output a specific format or hierarchical decoding, do not guarantee that the model follows the respective format. 
Furthermore, approaches such as hierarchical decoding of the respective ICD codes directly \cite{boyle_automated_2023} do not make use of the textual generative properties of the LLM. 

\begin{figure*}[t!]
    \centering
    \includegraphics[width=\linewidth]{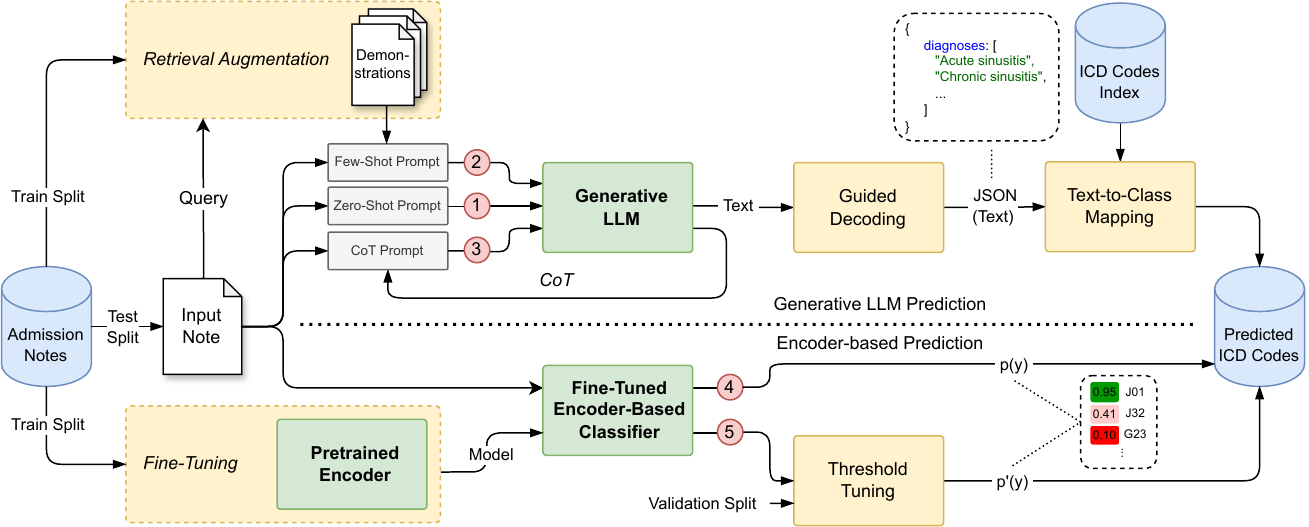}
    \caption{Architecture diagram of the CliniBench benchmark. Depicted are both model classes: Generative LLM (top) and encoder-based classifier (bottom). In addition to zero-shot evaluation (1), the LLM can be augmented via retrieval augmentation (2) and chain-of-thought (CoT) prompting (3) and predicts text (valid JSON) that is mapped to distinct labels. The encoder (bottom) predicts a probability distribution over the possible diagnoses (4) which can be threshold-tuned to improve performance (5).}
    \label{fig:framework}
\end{figure*}

\section{CliniBench}
In order to compare encoder-based classifiers and generative LLMs we build the CliniBench benchmark depicted in \cref{fig:framework}. The benchmark consists of the task of predicting a patient's outcome in the form of discharge codes for a clinical note at admission.

\subsection{MIMIC-IV Dataset}

The source of medical texts is the Medical Information Mart for Intensive Care (MIMIC-IV) dataset. MIMIC-IV is grouped into two modules: HOSP and ICU. Data in the HOSP module are sourced from the hospital-wide electronic health record (EHR), while data in the ICU module are sourced from the clinical information system from the intensive care unit (ICU).

\paragraph{Admission Notes.} To support clinical decision-making through diagnosis prediction, we limit the extracted information to relevant data known at admission time. We exclude sections containing outcome-related information obtained during the stay or after hospitalization, and exclude administrative or demographic fields that are de-identified or do not contribute to the diagnosis prediction task (see \cref{sec:appendix:admission_note_sections} for complete section details).

\paragraph{ICD Discharge Codes.} Both modules include detailed discharge codes derived from the International Classification of Diseases (ICD) coding system, documenting diseases and medical conditions (morbidity) associated with each admission. Depending on the date of admission, codes are available in either ICD-9 or ICD-10 formats (see \cref{table:dataset-stats}). Given the hierarchical nature of the ICD system, we reduce the granularity of each code to three digits, allowing encoder-based models to learn from multiple examples rather than single instances during fine-tuning. This reduces the average number of codes per note to 13.98 while still providing adequate clinical information.

\paragraph{Dataset Construction.}
We derive CliniBench dataset from the MIMIC-IV-Note dataset \cite{johnson_mimic-iv-note_v2_2_2023} and ICD discharge codes from the MIMIC-IV dataset \cite{johnson_mimic-iv_v2_2_2023} following \citet{van-aken-etal-2021-clinical-outcome}.
Each instance corresponds to one hospital admission note (see Appendix \ref{sec:appendix:admission_note_sections} for construction details).

\paragraph{Dataset Split.}
We generate a stratified split into training (70\%), validation (10\%), and test (20\%) sets.
The dataset comprises four partitions reflecting coding system and care setting: ICD-9/ICD-10 × HOSP/ICU.
We exclude labels not present across all three splits from evaluation.
We additionally report the notes-per-label cutoffs used to define frequency tertiles: Tail contains labels with frequency $\leq t_{tail}$, Head contains labels with frequency $> t_{head}$ and Body contains labels with $t_{tail} > freq \leq t_{head}$. Cutoffs are computed on the train split per partition.
Detailed dataset statistics are provided in Table \ref{table:dataset-stats}.

\begin{table}
\resizebox{\linewidth}{!}{%
\centering

\begin{tabular}{lrrrr}
\toprule
                          & \multicolumn{2}{c}{\textbf{ICU}} & \multicolumn{2}{c}{\textbf{HOSP}} \\
                          & ICD-9 & ICD-10 & ICD-9 & ICD-10 \\
\midrule
\textbf{\#Train Adm. Notes} & 27007 & 18596 & 119025 & 66736 \\
\textbf{\#Val Adm. Notes}   & 3618   & 2399   & 15683   & 8912 \\
\textbf{\#Test Adm. Notes}  & 8089  & 5600  & 35930  & 20058 \\
\midrule
\textbf{\#Full Codes}         & 6436 & 9610 & 8209 & 14362 \\
\textbf{\#Short Codes}         & 1038 & 1472 & 1105 & 1616 \\
\midrule
\textbf{Avg. \#Tokens/Adm. Note}  & 670.4 & 624.9 & 681.5 & 687.5 \\
\textbf{Avg. \#Codes/Adm. Note} & 14.6 & 18.4 & 10.0 & 12.9 \\
\midrule
\textbf{\#Adm. Notes/Code (Tail) $\le$} & 8  & 7  & 17  & 10 \\
\textbf{\#Adm. Notes/Code (Head) $>$}   & 55 & 43 & 126 & 67 \\
\bottomrule
\end{tabular}
}
\caption{
Statistics of the four CliniBench partitions derived from MIMIC-IV admission notes.
We report train/validation/test sizes, label set sizes, average label and token counts, and admission notes-per-label cutoffs for the frequency tertiles.
Token counts are computed using the BioMedBERT tokenizer.
}
\label{table:dataset-stats}
\end{table}

\subsection{Model Classes and Augmentation}
\cref{fig:framework} illustrates the process of predicting ICD codes from admission notes using two primary approaches: A generative, LLM-based approach and an encoder-based approach. 


\paragraph{Encoder.}
For the encoder-based prediction pipeline, we fine-tune a pretrained transformer to predict ICD diagnosis code probabilities.
To ensure comparability with generative models, we evaluate using threshold-based metrics like recall and precision.
Since thresholds can be noisy for rare classes, we tune them by maximizing the $F_1$ score on the Precision-Recall curve. We report details on the threshold-tuning in Section \ref{experiment_encoder}.

\paragraph{Generative LLMs.}

For the generative LLMs, we prompt each model to output a list of short diagnosis descriptions. 
We apply guided decoding \cite{guided_generation-louf-2023} to ensure a machine-readable JSON format. 
These structured outputs are mapped to corresponding ICD codes using text-based matching.
We investigate three distinct approaches: Zero-Shot, Few-Shot, and Zero-Shot with CoT prompting.

\paragraph{Zero-Shot.} In the Zero-Shot setting, we input the admission notes with task-specific instructions.
Thus, the model can only leverage knowledge acquired from its initial pretraining and the information available in the shown example.

\paragraph{Few-Shot.} 
Generative LLMs can learn from few-shot examples provided in the prompt \cite{few_shot_gpt3, song-2023-survey-few-shot-learning, parnami-2022-summary-few-shot-learning} without adapting the parameters (in-context-learning).
For this adaptation, either the user or a system incorporates complementary knowledge in the prompt. 
This is especially useful for domain specific tasks that are not well represented in the pretraining data.

\paragraph{Chain-of-Thought.} 
Finally, we apply chain-of-thought reasoning by prompting the model to produce reasoning steps prior to generating the list of diagnoses.
We hypothesize that incorporating reasoning can improve the accuracy and diversity of predictions by mirroring the process of differential diagnosis in real-world clinical settings.
\section{Experimental Setup}

\subsection{Encoder}
\label{experiment_encoder}

Previous work has shown that encoders in the clinical setup benefit from in-domain pretraining \cite{lee-2020-biobert}.
Therefore, we use two encoder models that are fine-tuned for clinical tasks: BioMedBert~\cite{gu_domain-specific_2021} and GatorTronS~\cite{Peng2023-sm}.
Since preliminary results show a decline of model performance beyond their context length (see \cref{fig:token_count_reg}), we additionally evaluate a state-of-the-art long context encoder model, M2-BERT~\cite{saad-falcon_benchmarking_2024}.

\paragraph{Fine-tuning and hyperparameter optimization.}
We fine-tune the pretrained encoders on the train split and select the best performing using AUROC on the validation split using AdamW \cite{Loshchilov2019AdamW} as optimizer and a polynomial learning rate decay. For the hyperparameter optimization, we apply Bayesian optimization. For more details compare \cref{table:hpo-params} and \cref{table:auroc}. 

\paragraph{Threshold tuning.}
We optionally calibrate encoder outputs via class-wise threshold tuning on the validation split.
For each label $c$, we evaluate $F_1$ over the sorted candidate threshold list $T_c$ and choose $j=\arg\max F_c$; we set $\tau_c=(T_c[j-1]+T_c[j])/2$ (or $\tau_c=T_c[0]$ if $j=0$).
This midpoint rule avoids a fixed offset when thresholds are unevenly spaced.
After calibration we still select the top-$n$ labels by score; tuning only changes label ordering, primarily affecting low-frequency labels.

\subsection{Generative LLMs}
We evaluate open-weight generative LLMs that are capable of processing sequences exceeding 4096 tokens. This extended sequence capacity is crucial for evaluating models in the few-shot setting, as sequence length increases with each added demonstration. 
Consequently, we excluded both available medical LLMs MedAlpaca and Med42 from our analysis, as both are based on the Llama2 architecture, which limits their maximum sequence length to 2048 tokens.
We selected the following models for evaluation: Llama 3.1 8B and 70B, Mistral 7B, Mistral Nemo (13B), Qwen2 7B and 72B.
For each model, we evaluate both the instruction fine-tuned and the non-instruction-tuned variants, resulting in a total of 12 models.
We explicitly do not benchmark closed-source commercial models due to constraints of the Physionet license \footnote{\url{https://physionet.org/content/mimiciv/view-license/2.2/}}.

%

\paragraph{Guided decoding.}
While generative models often produce valid JSON outputs and the correct number of prompted diagnoses, some of the evaluated models, especially those not fine-tuned with instruction data, frequently fail to do so. This limitation reduces overall task performance compared to encoder-based models.
We report the performance of generating valid JSON output for all models in the Appendix in \cref{table:json-performance}.
To mitigate this issue, we apply guided decoding \cite{guided_generation-louf-2023} to enforce JSON output that contains exactly 20 strings. Each string represents a diagnosis description between 3 and 70 characters in length. To optimize performance, we limit the number of generated tokens to a maximum of 1500. Finally, we use greedy sampling with a temperature of 0.

\paragraph{Prompts.}
We use multiple prompts for each experiment: Zero-shot, zero-shot with CoT, and few-shot learning. The prompts consist each of a role description for the model and the respective task, further detailed in \cref{sec:prompts}.

\paragraph{Text-to-class mapping.}
Because LLMs often struggle to predict exact ICD codes \cite{boyle_automated_2023}, we instead generate diagnostic descriptions and map them to ICD codes via their corresponding long or short descriptions. 
Based on a comparison of different approaches, we adopt NeuML-PubMedBERT\footnote{\href{https://huggingface.co/NeuML/pubmedbert-base-embeddings}{https://huggingface.co/NeuML/pubmedbert-base-embeddings}} for text-to-label mapping. Details are reported in the Appendix in \cref{table:class-mapper-performance}. 
Finally, we truncate ICD codes to three digits to reduce label space and remove duplicates from the predictions.

\subsection{Retrieval Models and Few-Shot Learning}
We implement few-shot learning by incorporating one, three, or five demonstrations into the prompt. 
The effectiveness of in-context-learning  depends on the selection of demonstrations \cite{liu-2022-icl-analysis}.
Thus, we use a retriever to find the most relevant candidates.
The retriever takes one admission note as input query and returns the most similar admission notes from the training split. 
We conduct three different retrieval strategies and compare them to \textit{random} and \textit{gold} demonstrations. Namely, we use document-based semantic similarity, latent outcome similarity (classifier fine-tuned on diagnosis prediction task), and a term-based retrieving strategy.

\paragraph{Retriever selection and majority voting.}
To assess retrieval model performance for clinical outcome prediction, we evaluate them directly on the diagnosis prediction task. We apply majority voting over five retrieved training documents, selecting the top 20 most frequent labels, prioritizing documents with higher similarity in case of ties. We evaluate several semantic similarity retrievers based on S-BERT \cite{sbert_reimers2019} (see \cref{appendix:retrieval-model-details} for complete list), selecting only the best performer for few-shot experiments to reduce computational costs. For latent outcome similarity retrieval, we use our fine-tuned BioMedBERT encoder (\cref{experiment_encoder}), hypothesizing that admission notes with similar outcomes cluster in embedding space. We also employ BM25 \cite{bm25_Robertson2009ThePR} for term-based retrieval.

\paragraph{Ideal retriever: \textit{gold heuristic}.}
\label{paragraph:gold}
To assess the potential benefits of few-shot learning, we simulate an ideal retriever.
Instead of retrieving based on semantic similarity, we sample demonstrations using the Otsuka similarity coefficient \cite{otsuka-similarity-1936} of annotated ICD codes.
Since this approach requires prior knowledge of the labels to identify similar patients, the experiment is designed solely to illustrate the upper limit of few-shot learning performance when leveraging similar patient examples.

\paragraph{Distractive demonstrations: \textit{random}.}
Finally, we assess random sampling to determine the extent to which the LLM relies on the provided demonstrations and whether irrelevant examples negatively impact the performance of the generative LLM. 
This approach also enables us to evaluate whether the model effectively adapts to the structure of the presented demonstrations.

\subsection{Evaluation Protocol and Metrics}

\paragraph{Fixed-size top-$n$ evaluation.}
We evaluate all methods in a fixed-size top-$n$ setting, where each model outputs exactly $n$ diagnosis codes per admission note.
A fixed output size is required to ensure comparability between encoder-based classifiers, which assign a score to each label, and generative LLMs, which produce an explicit list of codes.

For encoder models, we rank labels by predicted score and select the top-$n$ codes; when threshold tuning is applied, it affects score calibration and thus the ranking prior to top-$n$ selection.
For generative models, we constrain decoding to return a JSON list containing exactly $n$ codes, thus simplifying guided decoding schema and reducing the maximum number of generated tokens per instance.


We choose $n{=}20$ as a single global output budget that yields stable comparisons across partitions and supports a recall-oriented operating point.
This value is motivated by the observed label cardinality in CliniBench: across partitions, notes contain on average 10.0--18.4 short codes (Table~\ref{table:dataset-stats}), and our preliminary analysis indicates an $F_1$ plateau around this range when varying $n$.
We therefore use $n{=}20$ to avoid systematic under-prediction on high-cardinality notes while keeping the output budget fixed across model classes.
We treat $n$ as a benchmark hyperparameter that can be adjusted for different precision--recall preferences.

\paragraph{Evaluation Metrics.}
We report macro-averaged recall, precision, and $F_1$ as main metrics.
In addition, we report mean average precision (MAP), computed over the ranked list of predicted codes.
Finally, we report main diagnosis accuracy (MD Acc.), i.e., whether the first annotated diagnosis code is contained in the predicted top-$n$ list.

\paragraph{Upper ceiling metrics.}
A fixed top-$n$ protocol induces metric ceilings: for notes with fewer than $n$  codes, precision cannot reach 1.0; for notes with more than $n$ codes, recall cannot reach 1.0. These ceilings depend only on the set size $|Y|$, with $\mathrm{Prec}_{\max}(Y)=\min(|Y|,n)/n$ and $\mathrm{Rec}_{\max}(Y)=\min(|Y|,n)/|Y|$. On the test split with $n{=}20$, the resulting dataset-level ceilings are Prec.\ $=0.5566$ and Rec.\ $=0.9824$.
\section{Results}
\begin{table*}
	[h!]
	\centering
	\resizebox{\linewidth}{!}{
	\begin{tabular}{clrrrrr|rrrrr|rrrrr}
		\toprule                                         &                              & \multicolumn{5}{c|}{\textbf{Zero-Shot prompted}} & \multicolumn{5}{c|}{\textbf{Zero-Shot prompted + CoT}} & \multicolumn{5}{c}{\textbf{Few-Shot 5 Demonstrations}} \\
		\multicolumn{1}{l}{\textbf{Data}}                & \textbf{Model}               & \textbf{Rec.}                           & \textbf{Prec.}                                & \textbf{MAP}                                          & \textbf{MD Acc.}  & \textbf{F1}       & \textbf{Rec.}           & \textbf{Prec.}        & \textbf{MAP}  & \textbf{MD Acc.} & \textbf{F1}       & \textbf{Rec.}     & \textbf{Prec.} & \textbf{MAP}  & \textbf{MD Acc.} & \textbf{F1}       \\
		\midrule

\multirow{24}{*}{\rotatebox{90}{HOSP}} & \textbf{Encoder-based Models} \\
		                                                 & BioMedBERT-110M              & \textbf{30.74}                          & 19.91                                         & 33.61                                                 & 78.21             & 14.63             & \multicolumn{5}{c|}{--} & \multicolumn{5}{c}{--} \\
		                                                 & BioMedBERT-110M-tuned        & 22.91                                   & 29.55                                         & 19.34                                                 & 59.16             & 21.74             & \multicolumn{5}{c|}{--} & \multicolumn{5}{c}{--} \\
		                                                 & GatorTronS-345M              & 30.45                                   & 22.70                                         & \textbf{35.26}                                        & \textbf{80.94}    & 18.46             & \multicolumn{5}{c|}{--} & \multicolumn{5}{c}{--} \\
		                                                 & GatorTronS-345M-tuned        & 28.61                                   & \textbf{33.44}                                & 22.06                                                 & 65.22             & \textbf{25.96}    & \multicolumn{5}{c|}{--} & \multicolumn{5}{c}{--} \\
		                                                 & M2-80M                       & 14.90                                   & 08.51                                         & 24.34                                                 & 60.83             & 4.54              & \multicolumn{5}{c|}{--} & \multicolumn{5}{c}{--} \\
		                                                 & M2-80M-tuned                 & 11.72                                   & 12.05                                         & 14.80                                                 & 40.47             & 11.16             & \multicolumn{5}{c|}{--} & \multicolumn{5}{c}{--} \\
		\cmidrule(lr){2-17}                              & \textbf{Generative LLM}      &                                         &                                               &                                                       &                   &                   &                         &                       &               &                  &                   &                   &                &               &                  &                   \\[0.5ex]
		                                                 & Llama 3.1-8B                 & 16.61                                   & \underline{12.16}                             & 12.88                                                 & 35.22             & 11.06             & 14.85                   & 10.89                 & 11.49         & 37.96            & 09.68              & 21.76             & \uline{16.00}  & 21.33         & 47.23            & 12.41             \\
		                                                 & Llama 3.1-8B-Instruct        & 22.31                                   & 17.23                                         & 14.05                                                 & 47.53             & 11.61             & 20.84                   & 09.93                 & 11.95         & 49.61            & 10.11             & 20.99             & 14.24          & 18.06         & 44.34            & 15.45             \\
		                                                 & Llama 3.1-70B                & 17.29                                   & 09.78                                         & 11.40                                                 & 35.21             & 09.66             & 20.72                   & 09.42                 & 12.36         & 43.63            & 10.20             & 20.55             & 12.67          & 19.68         & 49.05            & 15.08             \\
		                                                 & Llama 3.1-70B-Instruct       & 23.04                                   & 10.40                                         & 12.48                                                 & 52.20             & 11.32             & 23.70                   & 10.00                 & 12.53         & 51.39            & 10.59             & 21.89             & 14.43          & 18.02         & 45.49            & 15.82             \\
		                                                 & Mistral-7B                   & 10.45                                   & 12.02                                         & 10.12                                                 & 27.25             & 08.50             & 00.40                   & 03.86                 & 02.29         & 00.69            & 00.15             & 19.15             & 13.78          & 18.02         & 44.13            & 15.29             \\
		                                                 & Mistral-7B-Instruct          & 26.45                                   & 10.52                                         & 14.88                                                 & \underline{54.06} & 12.16             & 23.96                   & 09.46                 & 12.81         & 49.76            & 10.86             & \underline{25.33} & 14.29          & 18.77         & 48.89            & 15.93             \\
		                                                 & Mistral-Nemo-13B             & 14.87                                   & 11.30                                         & 12.53                                                 & 32.83             & 10.15             & 13.66                   & 09.37                 & 09.90         & 38.32            & 08.00             & 19.18             & 13.76          & 18.44         & 45.97            & 15.10             \\
		                                                 & Mistral-Nemo-13B-Instruct    & 22.23                                   & 09.55                                         & 12.34                                                 & 50.35             & 10.38             & 21.19                   & 09.22                 & 10.44         & \uline{51.61}    & 09.57             & 21.15             & 15.28          & 19.09         & 48.41            & 16.05             \\
		                                                 & Qwen2-7B                     & 22.86                                   & 11.32                                         & 15.12                                                 & 43.37             & 12.26             & 21.04                   & \uline{11.62}         & 14.92         & 42.01            & 12.04             & 17.70             & 14.94          & 18.81         & 43.40            & 15.31             \\
		                                                 & Qwen2-7B-Instruct            & 25.80                                   & 11.27                                         & 15.28                                                 & 48.95             & 12.90             & 24.63                   & 10.76                 & 15.32         & 49.38            & 12.15             & 24.32             & 13.68          & 19.38         & 51.65            & 15.54             \\
		                                                 & Qwen2-72B                    & 24.63                                   & 11.06                                         & 15.45                                                 & 46.17             & 12.47             & 22.18                   & 10.42                 & 14.86         & 44.05            & 11.49             & 22.57             & 14.36          & 20.38         & 49.44            & 16.46             \\
		                                                 & Qwen2-72B-Instruct           & \underline{27.24}                       & 11.09                                         & \underline{17.28}                                     & 51.58             & \underline{13.13} & \uline{25.66}           & 10.91                 & \uline{15.28} & 50.24            & \underline{12.75} & 25.29             & 15.82          & \uline{21.51} & \uline{53.00}    & \underline{17.57} \\
		\bottomrule
	\end{tabular}
	}
	\caption{(\textbf{MAP}: \textit{Mean Average Precision}, \textbf{MD Acc.}: \textit{Main
	Diagnosis Accuracy}) Zero-shot performance on the HOSP split over the top-20 predictions.
	The reported metrics are macro averaged. We aggregate ICD-9 and ICD-10 and report
	an unweighted mean. CoT is not beneficial while instruct fine-tuning boosts
	performance. Threshold tuning decreases the performance of encoder models within
	the top-20 predictions. We depict the best performing models in bold and
	underline the best performing generative models in each category.}
	\label{table:big_results_table}
\end{table*}

\subsection{Zero-Shot}
Encoder-based architectures consistently outperform their generative counterparts with the exception of the M2 BERT model. 

\paragraph{Encoders} excel especially for predicting the main diagnosis with a main diagnosis accuracy (MD Acc.) of over 80\% on the HOSP split and 78.4\% on the ICU split.
When threshold tuning is applied, BioMedBERT and GatorTronS demonstrate strong performance decreases in recall, MD Acc., and MAP.
This effect can be explained by the fact that we only measure the performance within the first 20 of an average of 47.18 predicted labels.

\paragraph{Generative LLMs} benefit from instruction tuning, though they tend to lag behind encoder models. 
The addition of CoT reasoning yields inconsistent results for generative models, sometimes providing marginal improvements but often reducing MAP or recall.
We hypothesize that CoT induced reasoning could distract the model from the actual task. This aligns with results from other work which shows that CoT is only beneficial for tasks that involve mathematical or symbolic reasoning \cite{sprague-2024-cot-analysis}.

\subsection{Retrieval Models}
\cref{table:majority_voting_results} in the appendix shows the performance of retrieval models using majority voting on the diagnosis prediction task. 
We observe that our fine-tuned classifier, BioMedBERT$_{ft}$, achieves the best performance. 
Notably, the baseline BM25 outperforms all retrieval models based on semantic similarity. 
Among the sentence-transformer models, performance differences are minimal, with PubMedBERT$_{NeuML}$ outperforming others. 
These findings suggest that using domain specific retrieval models provides a modest advantage for clinical tasks.

\begin{table}[ht!]
\resizebox{\linewidth}{!}{%
\begin{tabular}{clrrrrr} 
\toprule
\textbf{Data}                & \textbf{Retriever}                          & \textbf{\#FS} & \textbf{Rec.} & \textbf{Prec.} & \textbf{MAP}   & \textbf{MD Acc.}  \\ 
\midrule
\multirow{16}{*}{\rotatebox{90}{HOSP}}
    & Zero-Shot prompted & 0 & \textbf{27.24} & 11.09 & 17.28 & 51.58 \\
    \cline{2-7}
    & \multirow{3}{*}{\textit{Random}}           & 1 & 21.53 & 11.77 & 15.14 & 42.66  \\
    &                                   & 3 & 20.14 & 12.25 & 15.88 & 42.86  \\
    &                                   & 5 & 18.90 & 12.32 & 17.75 & 41.59  \\ 
\cline{2-7}
    & \multirow{3}{*}{BM25}             
        & 1 & 25.12 & 13.97 & 19.33 & 49.62  \\
    &   & 3 & \uline{25.48} & \uline{15.17} & \uline{21.53} & 52.32  \\
    &   & 5 & 25.29 & \textbf{15.83} & \textbf{22.02} & \uline{53.00}  \\ 
\cline{2-7}
    & \multirow{3}{*}{PubmedBERT$_{NeuML}$} & 1 & 22.38 & 11.90 & 16.47 & 46.86  \\
    &                                   & 3 & 22.20 & 13.35 & 17.63 & 50.97  \\
    &                                   & 5 & 21.87 & 13.87 & 20.18 & 52.63  \\ 
\cline{2-7}
    & \multirow{3}{*}{BioMedBERT$_{ft}$}        & 1 & 23.05 & 12.59 & 16.94 & 46.59  \\
    &                                   & 3 & 23.52 & 13.89 & 20.35 & 51.24  \\
    &                                   & 5 & 24.25 & 14.63 & 20.89 & \textbf{53.04}  \\ 
\cline{2-7}
    & \multirow{3}{*}{\textit{Gold Heuristic}} & 1 & 27.04 & 15.80 & 20.97 & 51.36  \\
    &                       & 3 & 30.43 & 19.49 & 28.11 & 60.04  \\
    &                       & 5 & 32.04 & 20.77 & 31.90 & 62.62  \\
\bottomrule
\end{tabular}
}
\caption{Few-shot performance of Qwen2-72B-Instruct averaged over ICD-9 and ICD-10 splits using the depicted retrieval models. Metrics are macro averaged. BioMedBERT$_{ft}$ is a custom fine-tuned encoder model on the respective training split of the dataset. (\textbf{\#FS}: number of few-shot demonstrations, \textbf{MD Acc.}: Main diagnosis accuracy)}
\label{table:few_shot_results}
\end{table}

\subsection{Few-Shot} 

As shown in \cref{table:big_results_table}, \cref{table:few_shot_results} and \cref{fig:bar_few_shot} 
(ICU scores reported in Appendix \cref{table:big_results_table_icu} and \cref{table:few_shot_results_icu}), providing generative LLMs with even a single example enhances their performance compared to zero-shot settings. 
This improvement is likely because the demonstration helps align the model more closely with the task, particularly benefiting non-instruct variants of generative models. 
Adding more demonstrations continues to boost performance, albeit with diminishing gains.

\paragraph{Impact of retrieval quality.}
The \textit{gold} heuristic provides an idealized upper boundary for performance by retrieving the most relevant samples (see \cref{table:majority_voting_results}). 
It achieves the highest metrics across both splits, with the main diagnosis accuracy reaching 62.62\% for HOSP and 61.66\% for ICU, along with substantial MAP improvements. 
Among the retrievers, BM25 emerges as a robust baseline, consistently delivering competitive MAP and main diagnosis accuracy, outperforming NeuML-PubMedBERT and BioMedBERT$_{ft}$ in several configurations. 
Task-specific fine-tuning provides a slight advantage for BioMedBERT$_{ft}$, which outperforms NeuML-PubMedBERT.
Random retrieval degrades the performance of generative models, underscoring the importance of relevance-driven retrieval strategies for downstream tasks.


\paragraph{Performance across different label frequencies.}
\begin{figure}[h!]
    \centering
    \includegraphics[width=0.48\textwidth]{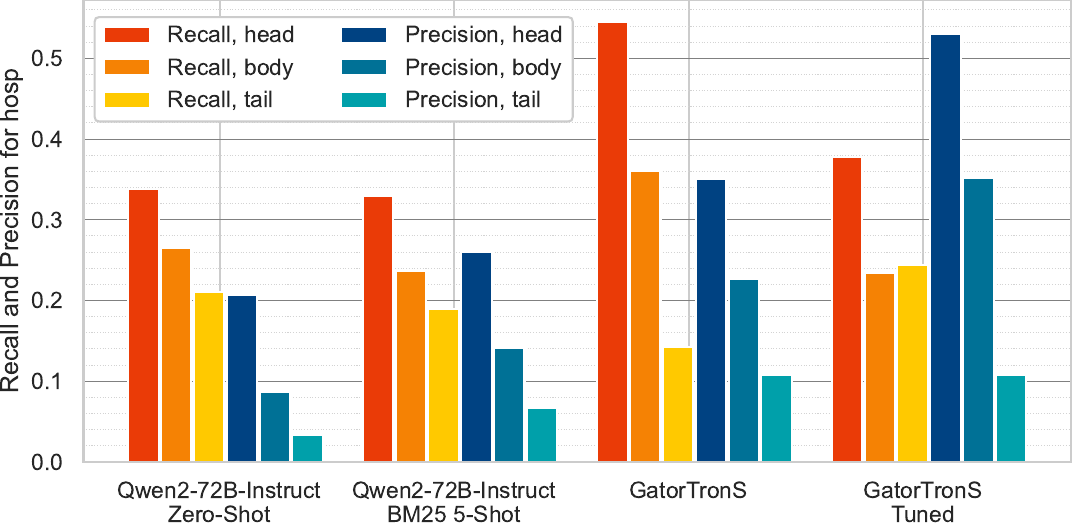}
    \caption{Macro recall and precision by model and classes grouped into tertiles by class frequency aggregated over both HOSP splits (ICD-9 and ICD-10).}
    \label{fig:bar_class_frequency}
\end{figure}

The results in \cref{fig:bar_class_frequency} (see also Appendix \cref{fig:bar_class_frequency_icu}) highlight variations in performance across labels with different frequencies.
We observe that general performance is better for frequent labels.
We find that retrieval augmentation boosts precision and reduces recall for all tertiles, particularly for less frequent labels.
Generative models achieve a higher recall for rare codes than encoder-based classifiers.
Finally, we observe that threshold tuning GatorTronS improves recall in the tail, while decreasing it in the head tertile.

\paragraph{Impact of model size.}
We observe that the number of parameters does not have a strong impact on performance.
Further, we find that few-shot learning especially benefits smaller models, increasing the performance drastically in contrast to larger models.
We hypothesize that guided decoding reduces the necessity of capacity in the larger models, making small models suitable for the task. 
Furthermore, we find that instruct fine-tuning has a larger effect on performance compared to the number of parameters (see \cref{fig:rad_recall}), narrowing the gap between small and large models.

\subsection{Error Analysis}
\label{sec:error_analysis}
\begin{figure}[htb!]
    \centering
    \includegraphics[width=0.48\textwidth]{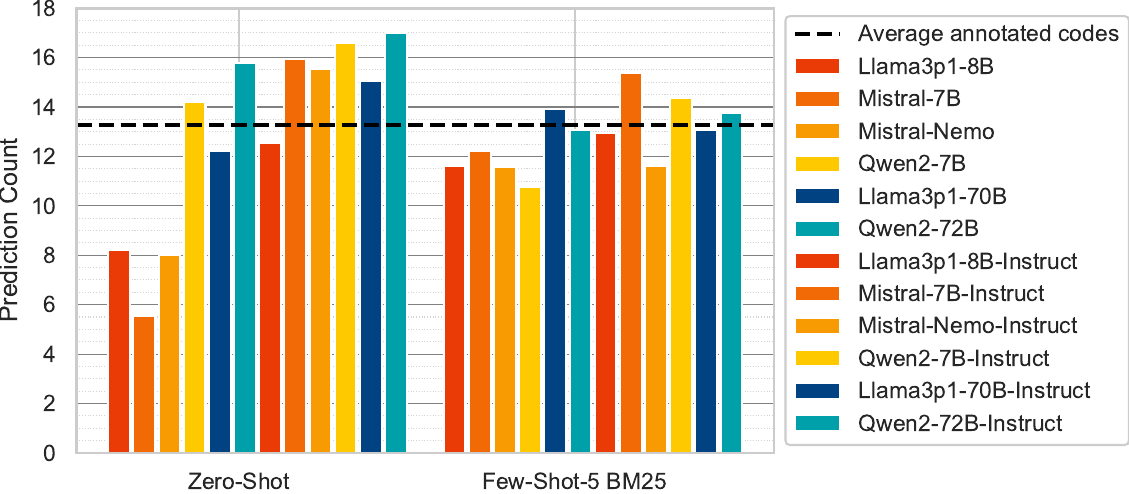}
    \caption{Average number of predicted codes after deduplication for zero-shot and few-shot experiments. Small LLMs are represented in shades of red, large ones in shades of blue, with instruction-tuned models highlighted in a shaded pattern.}
    \label{fig:pred_len}
\end{figure}

In addition to presenting quantitative results, we conduct a qualitative analysis of 20 randomly sampled examples with especially low macro recall from Qwen2-72B and Qwen2-72B-Instruct in zero-shot, few-shot (BM-25), and chain-of-thought settings. From our analysis, we deduce the following error classes:

\paragraph{Low variance outputs in generative LLMs.}
We observe that generative models sometimes generate text with little to no variance (e.g., \textit{Lung Cancer, Lung Carcinoma, Cancer}), leading to fewer predicted codes after deduplication. This behavior appears mainly in few-shot augmentation (80\% of one-shot, 75\% of five-shot predictions) and 50\% of CoT predictions. As shown in \cref{fig:pred_len}, all models predict fewer than the expected 20 codes on average, suggesting this issue exists across scenarios. Non-instruct fine-tuned models generate more varied output and thus more diverse diagnoses.

\paragraph{Unrelated generated text.}
We find that LLMs frequently produce valid but irrelevant text (e.g., procedures instead of diagnoses) or non-meaningful text (e.g., \textit{diagnosis 1, diagnosis 2, diagnosis 3}). In zero-shot settings with non-instruct models, 65\% of errors originate from unrelated meaningful output, where models often output procedures or copy parts of the admission note. Interestingly, a single few-shot example increases non-meaningful text generation in 20\% of cases, suggesting the model focuses on structural rather than task-specific elements.


\section{Future Work}
To address the limitations identified in our study, we propose several directions for future research that could substantially improve the performance of generative models for clinical outcome prediction.

\paragraph{Output Verification and Sampling}
To address low-variance and unrelated outputs, we suggest adding LLM-based verification mechanisms \cite{gero_self-verification_2023} to filter predictions and reduce duplicates. Additionally, probabilistic sampling strategies (top-p, top-k, beam search) could increase output diversity.

\paragraph{Enhanced Retrieval Strategies}
Given that latent outcome similarity outperforms semantic similarity and the gold heuristic surpasses both, future work should augment retrieval with additional clinical information beyond admission notes, such as laboratory results or imaging data, to improve demonstration selection for few-shot learning.

\paragraph{Domain Adaptation and Continual Learning}
To enable effective chain-of-thought reasoning, we suggest adapting models using clinical guidelines and datasets that capture clinicians' differential diagnosis processes \cite{wang2025directdiagnosticreasoningclinical}. Future research should also explore efficient fine-tuning strategies that balance performance with computational costs for scenarios requiring frequent model updates in response to evolving medical knowledge.

\section{Conclusion}


This study introduces CliniBench, a benchmark designed to comprehensively evaluate and compare generative LLMs and encoder-based classifiers for diagnosis prediction.
Our findings underscore that the encoders maintain superior performance.
However, generative LLMs exhibit promising adaptability through in-context learning strategies which are advantageous in resource-constrained or rapidly evolving environments such as the medical domain.
CliniBench provides a systematic framework to analyze the limitations of generative models, particularly high inference costs, lack of output diversity, and susceptibility to produce unrelated output.
By highlighting necessary improvements in retrieval strategies, sampling techniques, and domain-specific model adaptations, CliniBench paves the way for future research to bridge the performance gap and unlock the full potential of generative LLMs in clinical decision support systems.

\section{Acknowledgements}
We thank the reviewers and the Area Chair for their constructive feedback, which improved this work. This work is funded by the Deutsche Forschungsgemeinschaft (DFG, German Research Foundation) ProjectID 528483508- FIP 12, as well as the European Union under the grant project 101079894 (COMFORT- Improving Urologic Cancer Care with Artificial Intelligence Solutions), as well as by BMWe SOOFI, Grant Agreement 13IPC040D. The views expressed are solely those of the authors and do not necessarily reflect those of the European Union or European Health and Digital Executive Agency (HaDEA); neither is responsible for them.

\section{Limitations}
Our findings should be interpreted within the context of several study-specific limitations. 
First, our experiments rely on a single large-scale clinical dataset of admission notes, which may not represent the diversity of healthcare systems, patient populations, or note-taking styles.
Second, the evaluation focuses on diagnosis prediction from admission-time text and does not assess other clinically relevant outcomes (e.g., treatment recommendations or prognostic reasoning).
Third, admission notes frequently exceed the context length of standard encoder models; for encoders with limited attention window, we truncate inputs to the first 512 tokens, which may omit clinically relevant information and bias performance toward evidence appearing early in the document. We partially mitigate this by additionally evaluating a long-context encoder (M2-BERT), but a systematic comparison of long-document strategies (e.g., chunking or hierarchical encoding) is outside the scope of this benchmark.
Furthermore, we restrict our evaluation to open-weight models that can be executed locally. Due to the PhysioNet credentialed data use agreement for MIMIC-IV, we do not send protected note text to third-party commercial APIs and therefore do not benchmark closed-source models. As a result, our comparisons do not cover proprietary systems.
Finally, while we evaluate multiple architectures and prompting strategies, our hyperparameter choices and model selection may not reflect optimal configurations for every method. Future work should explore broader tuning budgets and additional modeling choices to assess the robustness and generalizability of our conclusions.

\section{Ethical Considerations}
The use of language models for diagnosis prediction raises important ethical concerns. For example, inaccurate or biased predictions can negatively impact patient care, as admission notes may lack critical information. It is important to note that an automatic diagnosis prediction system should never be seen as a replacement for medical personnel. Rather, it should be used as a supporting tool in the differential diagnosis process. Another major ethical concern is data privacy, as third-party data access can lead to serious consequences, including financial harm, discrimination, and compromised care. Finally, there is a risk that ICD code prediction might be misused to maximize profits, prioritizing financial incentives over patient care. Addressing these concerns is vital for the safe and effective use of AI in healthcare in the future.

\bibliography{tacl2021}

\appendix

\appendix
\section{Appendix}


\subsection{Prompts}
\label{sec:prompts}
For few-shot demonstrations, we provide the admission notes together with their annotated diagnosis short descriptions in the desired JSON format. 
We use the following prompt structure for each respective task setting:

\paragraph{Zero-shot:} "You are a medical professional. Given an admission note for a patient, present a list of possible diagnoses for the patient. The admission note is as follows: \{note\}."
\paragraph{Zero-shot with CoT:} "You are a medical professional. Given an admission note for a patient, present a list of possible diagnoses for the patient. The admission note is as follows: \{note\}. TASK: Solve this task step by step and give an explanation in maximum one or two sentences for each diagnosis decision."
\paragraph{Few-shot:} "You are a medical professional. Given an admission note for a patient, present a list of possible diagnoses for the patient. Similar patients look like this: \{few\_shots.json()\}. The admission note is as follows: \{note\}. Give the diagnoses following the schema from the examples."

We hypothesize that providing the model with comprehensive information about the ICD ontology would enhance its performance. However, the full set of code descriptions exceeds the maximum context length of all evaluated models. Using the Qwen2.5 tokenizer, the descriptions of all possible ICD-9 codes require 167,679 tokens, and those of ICD-10 codes require 1,369,475 tokens. Consequently, we restrict the input to the task prompt, admission note and few-shot examples.

\paragraph{CoT implementation.} 
For chain-of-thought reasoning, we modify the prompt to require the model to articulate specific reasoning steps before generating the list of diagnoses descriptions. Additionally, we adapt the expected JSON schema to include a reasoning field preceding the list of diagnoses. This ensures that the model first produces reasoning steps, followed by the diagnosis descriptions.

\subsection{Construction of the Admission Notes}
\label{sec:appendix:admission_note_sections}
To avoid data leakage and maintain temporal consistency, we carefully select which sections of clinical notes to include based on whether the information would realistically be available at admission time.

Included sections with information available at admission time:
\textit{Chief Complaint, Major Surgical or Invasive Procedure, Allergies, History of Present Illness, Past Medical History, Social History, Family History, Physical Exam at Admission, and Medication at Admission}

Discarded sections that contain information not available at admission time:
\textit{Physical Exam during Stay and at Discharge, Pertinent Results, Brief Hospital Course, Medication at Discharge, Discharge Disposition, Facility, Discharge Diagnosis, Discharge Condition, Discharge Instructions, and Followup Instructions}

Sections that do not contribute to the task or prediction and are therefore discarded:
\textit{Name, Unit No, Admission Date, Discharge Date, Date of Birth, Sex, Service, and Attending.
This systematic approach ensures that our models are evaluated under realistic clinical conditions where only admission-time information is available for outcome prediction}

\subsection{HPO and Threshold Tuning}

See \cref{table:hpo-params} for a breakdown of our hyperparameter optimization.



\begin{table*}[htb]
\centering
\resizebox{\linewidth}{!}{%
\begin{tabular}{ccl!{\vrule width \lightrulewidth}rrrrrc} 
\toprule
\multicolumn{3}{c!{\vrule width \lightrulewidth}}{\textbf{Parameter }} & \multicolumn{1}{c}{\textbf{Learning Rate}} & \multicolumn{1}{c}{\textbf{Warmup Steps}} & \multicolumn{1}{c}{\textbf{Weight Decay}} & \multicolumn{1}{c}{\textbf{Decay Steps}} & \multicolumn{1}{c}{\textbf{Batch Size}} & \textbf{Mean Pooling}                \\ 
\midrule
\multicolumn{3}{c!{\vrule width \lightrulewidth}}{\textbf{Ranges }}    & $[1\mathrm{e}{-6}, 1\mathrm{e}{-3}]$       & $[0, 5000]$                               & $[1\mathrm{e}{-6}, 1\mathrm{e}{-3}]$      & $[1,2,..7] * 10000$                      & $[8, 64]$                               & \multicolumn{1}{r}{$[False, True]$}  \\ 
\midrule
\multirow{4}{*}{BioMedBERT} & \multirow{2}{*}{HOSP} & ICD-9            & 0.000131                                   & 994                                       & 0.026853                                  & 50000                                    & 32                                      & False                                \\
    &                       & ICD-10           & 0.000526                                   & 146                                       & 0.095573                                  & 10000                                    & 64                                      & False                                \\
    & \multirow{2}{*}{ICU}  & ICD-9            & 0.000305                                   & 28                                        & 0.020785                                  & 10000                                    & 64                                      & False                                \\
    &                       & ICD-10           & 0.000158                                   & 374                                       & 0.008371                                  & 50000                                    & 16                                      & False                                \\ 
\midrule
\multirow{4}{*}{GatorTronS} & \multirow{2}{*}{HOSP} & ICD-9            & 0.000086                                   & 2185                                      & 0.034802                                  & 50000                                    & 16                                      & False                                \\
    &                       & ICD-10           & 0.000141                                   & 665                                       & 0.002609                                  & 40000                                    & 16                                      & False                                \\
    & \multirow{2}{*}{ICU}  & ICD-9            & 0.000098                                   & 4854                                      & 0.039681                                  & 50000                                    & 16                                      & False                                \\
    &                       & ICD-10           & 0.000091                                   & 4691                                      & 0.001923                                  & 40000                                    & 16                                      & False                                \\ 
\midrule
\multirow{4}{*}{M2-BERT}    & \multirow{2}{*}{HOSP} & ICD-9            & 0.000119                                   & 1728                                      & 0.000390                                  & 30000                                    & 8                                       & False                                \\
    &                       & ICD-10           & 0.000029                                   & 1698                                      & 0.000003                                  & 70000                                    & 8                                       & False                                \\
    & \multirow{2}{*}{ICU}  & ICD-9            & 0.000121                                   & 2376                                      & 0.000001                                  & 50000                                    & 16                                      & False                                \\
    &                       & ICD-10           & 0.000244                                   & 3620                                      & 0.000001                                  & 60000                                    & 16                                      & True                                 \\
\bottomrule
\end{tabular}
}
\caption{Hyperparameter ranges and best found hyperparameters for the evaluated encoder models. We perform 20 HPO runs using Bayesian optimization for each combination of model and dataset.}
\label{table:hpo-params}
\end{table*}

\subsection{Retrieval Model Details}
\label{appendix:retrieval-model-details}
The retrieval evaluation enables us to gauge the impact of adding a generative LLM on top of the retrieved few-shot results. We evaluate the following semantic similarity retrievers based on the S-BERT architecture \cite{sbert_reimers2019}: NeuML-PubmedBERT, jina-base-v2 \cite{günther-2023-jina}, all-mpnet-base-v2\footnote{\href{https://huggingface.co/sentence-transformers/all-mpnet-base-v2}{https://huggingface.co/sentence-transformers/all-mpnet-base-v2}}, all-mpnet-base-v2-neg \cite{anschutz-etal-2023-negation}, BiomedBERT \cite{gu_domain-specific_2021}, and S-PubMedBert-MS-MARCO \cite{deka-2022-s-pubmedbert-ms-marco}.
For the latent outcome similarity retrieval, we use the CLS token from the last embedding layer of our fine-tuned baseline encoder (BioMedBERT) as a representation of the admission note.

\begin{table}[htb!]
	\resizebox{\linewidth}{!}{%
		\begin{tabular}{llrrrr} 
			\toprule
			\textbf{Data}          & \textbf{Model}            & \textbf{Rec.}  & \textbf{Prec.} & \textbf{MAP}   & \textbf{MD Acc.} \\ 
			\midrule
			\multirow{11}{*}{HOSP} & BM25                      & \uline{14.73}  & \uline{8.73}   & \uline{16.97}  & \uline{45.22}    \\
			                       & BioMedBERT$_{ft}$         & \textbf{18.36} & \textbf{9.97}  & \textbf{19.36} & \textbf{49.16}   \\
			                       & BioMedBERT                & 07.07          & 03.98           & 12.52          & 37.43            \\
			                       & all-mpnet-base-v2         & 12.86          & 07.58           & 14.74          & 42.55            \\
			                       & all-mpnet-base-v2-neg     & 10.75          & 06.59           & 13.82          & 40.59            \\
			                       & jina-base-v2-512          & 13.42          & 08.05           & 15.33          & 43.61            \\
			                       & jina-base-v2-2048         & 12.69          & 08.16           & 15.51          & 43.80             \\
			                       & PubMedBERT$_{NeuML}$      & 13.57          & 08.13           & 15.89          & 44.25            \\
			                       & S-PubMedBert$_{MS-MARCO}$ & 11.64          & 07.40            & 14.84          & 42.82            \\ 
			                       & M2-BERT                   & 02.67          & 01.45           & 08.15           & 26.23            \\
			                       & GatorTronS                & 02.87          & 01.38           & 09.20            & 21.27            \\
			\cdashline{2-6}
			                       & \textit{Random}           & 01.96           & 01.06           & 07.04           & 23.66            \\
			                       & \textit{Gold  Heuristic}  & 31.67          & 23.45          & 40.93          & 73.36            \\ 
			\cmidrule{1-6}
			\multirow{11}{*}{ICU}  & BM25                      & \uline{09.57}   & \uline{09.11}  & 18.39  & 42.25            \\
			                       & BioMedBERT$_{ft}$         & \textbf{11.73} & \textbf{09.92}  & \textbf{19.99} & \textbf{45.08}   \\
			                       & BioMedBERT                & 05.11           & 04.63           & 15.27          & 36.88            \\
			                       & all-mpnet-base-v2         & 08.45           & 07.94           & 17.47          & 42.10             \\
			                       & all-mpnet-base-v2-neg     & 07.16           & 07.14           & 16.47          & 39.29            \\
			                       & jina-base-v2-512          & 08.38           & 08.23           & 17.89          & 41.96            \\
			                       & jina-base-v2-2048         & 08.04           & 07.99           & 17.91          & 42.18            \\
			                       & PubMedBERT$_{NeuML}$      & 09.12           & 08.40            & \uline{18.61}          & \uline{42.69}    \\
			                       & S-PubMedBert$_{MS-MARCO}$ & 07.82           & 07.65           & 17.56          & 41.24            \\ 
			                       & M2-BERT                   & 02.93           & 02.20            & 11.71          & 28.03            \\
			                       & GatorTronS                & 03.34          & 02.73           & 12.81          & 25.54            \\
			\cdashline{2-6}
			                       & \textit{Random}           & 02.47           & 01.90            & 10.49          & 26.14            \\
			                       & \textit{Gold  Heuristic}  & 19.98          & 19.85          & 37.69          & 64.41            \\
			\bottomrule
		\end{tabular}
	}
\caption{Independent analysis of retrieval models on the diagnosis prediction task. Metrics are macro averaged and the results are averaged over the respective ICD-9 and ICD-10 splits. A fine-tuned encoder works best (bold), followed by BM25 (underlined). The domain adapted PubMedBERT$_{NeuML}$ performs best from the sentence transformer models.}
\label{table:majority_voting_results}
\end{table}

\subsection{Computational Cost Analysis}

\begin{table}
\centering

\resizebox{\linewidth}{!}{%
\begin{tabular}{lr|rr|rr} 
\toprule
   & & \multicolumn{2}{c|}{\textbf{ICU}} & \multicolumn{2}{c}{\textbf{HOSP }}  \\
\textbf{Model}        &       \textbf{\#Parameters}                 & \textbf{ICD 9} & \textbf{ICD 10}  & \textbf{ICD 9} & \textbf{ICD 10}    \\ 
\midrule
BioMedBERT       & 110M                  & 1.17           & 0.73             & 5.23           & 2.89               \\
GatorTronS       & 345M                  & 3.45           & 2.16             & 15.41          & 8.53               \\
M2 BERT          & 80M                   & 1.12           & 0.72           & 4.97           & 2.80                  \\
Llama 3.1 8B     & 8B                    & 103.08         & 64.68            & 454.58         & 252.1              \\
Llama 3.1 70B    & 70B                   & 901.93         & 565.97           & 3977.57        & 2206.30            \\
Mistral 7B       & 7B                    & 109.79         & 68.84            & 484.08         & 267.82             \\
Mistral Nemo 13B & 13B                   & 181.99         & 113.66           & 802.34         & 441.28             \\
Qwen 2 7B        & 7B                    & 93.15          & 58.27            & 409.40         & 226.76             \\
Qwen 2 72B       & 72B                   & 599.37         & 958.10           & 2332.40        & 4210.96            \\
\bottomrule
\end{tabular}
}
\caption{Inference cost in PFLOPs for each test split using the respective tokenizer of the model. We approximate the number of FLOPs based on \cite{kaplan-scaling-laws-2020} by multiplying the number of tokens of each dataset with the number of parameters of the model multiplied by 2. Qwen2, due to its more efficient tokenizer, reduces compute costs compared to Llama 3.1 by almost 40\%.}
\label{table:flops}

\end{table}
The computational costs differ dramatically between model types. Training a BERT-base model on the ICD-10 Hosp dataset requires about 9.51 PFLOPs per epoch \cite{kaplan-scaling-laws-2020}, or 28.53 PFLOPs when including the backward pass. With a maximum of 20 training epochs, this totals 570.6 PFLOPs for our experiments.
By contrast, simply evaluating the ICD-10 Hosp test set with generative models is far more expensive: Qwen2-72B requires approximately 4210.96 PFLOPs, while Llama 3.1 8B needs 252.1 PFLOPs. The smaller BERT-based model only uses only 2.89 PFLOPs for inference on the same test set (see \cref{table:flops}). This demonstrates that generative models require orders of magnitude more computation than encoder-based models, even for inference on a small test set.
These estimates ignore several real-world considerations, such as infrastructure requirements and hardware utilization. Larger generative models demand substantially more hardware during inference, which might remain underutilized between tasks, though these costs could be partially offset by leveraging techniques such as pruning and quantization.

\clearpage
\newpage

\section{Additional Results}

\begin{table}[htb]
\scriptsize
\centering
\begin{tabular}{lllr} 
\toprule
\multicolumn{1}{l}{\textbf{Model}} & \multicolumn{2}{l}{\textbf{Data}} & \multicolumn{1}{c}{\textbf{AUROC}}  \\
\midrule
\multirow{4}{*}{BioMedBERT}         & \multirow{2}{*}{HOSP      } & ICD-9   & 90.12                               \\
    &                       & ICD-10  & 88.01                               \\
    & \multirow{2}{*}{ICU}  & ICD-9   & 83.59                               \\
    &                       & ICD-10  & 80.38                               \\ 
\midrule
\multirow{4}{*}{GatorTronS}         & \multirow{2}{*}{HOSP} & ICD-9   & 91.96                               \\
    &                       & ICD-10  & 89.43                               \\
    & \multirow{2}{*}{ICU}  & ICD-9   & 86.23                               \\
    &                       & ICD-10  & 82.17                               \\ 
\midrule
\multirow{4}{*}{M2-BERT}            & \multirow{2}{*}{HOSP} & ICD-9   & 85.82                               \\
    &                       & ICD-10  & 81.05                               \\
    & \multirow{2}{*}{ICU}  & ICD-9   & 77.99                               \\
    &                       & ICD-10  & 75.08                               \\
\bottomrule
\end{tabular}
\caption{AUROC performance of the fine-tuned encoder models.}
\label{table:auroc}
\end{table}

\begin{figure}[h!]
    \centering
    \includegraphics[width=\linewidth]{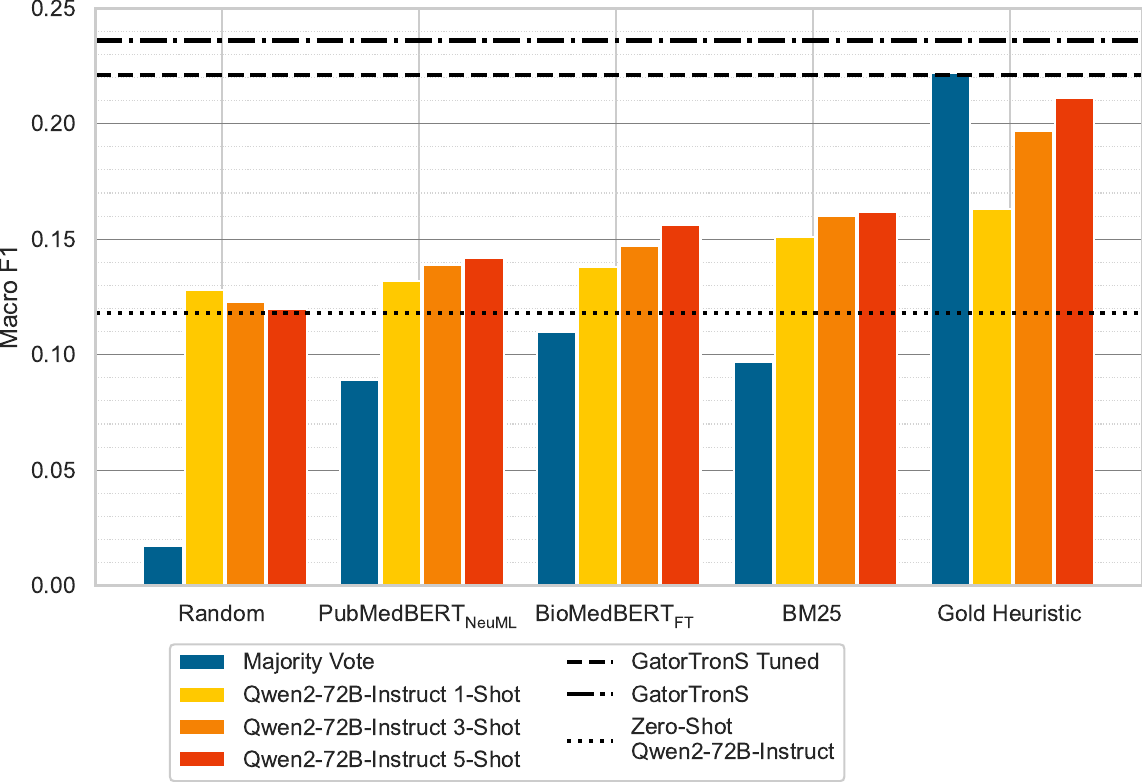}
    \caption{Macro $F_1$ (averaged over all datasets) of majority voting, best encoder, and best generative LLM with various amounts of demonstrations. Categories on the x-axis show different retrieving strategies for few-shot experiments. Horizontal lines depict the performance of the best zero-shot generative model and the best encoder models.}
    \label{fig:bar_few_shot}
\end{figure}

\subsection{Text-to-class Mapper Performance.}

We find that generative models often output diagnostic descriptions that exactly match the short or long descriptions of the respective ICD codes. 
This allows us to independently compare the performance of exact matched descriptions and codes that are mapped via the class mapper shown in \cref{table:class-mapper-performance}.
For Qwen2-72B Instruct in the zero-shot scenario, only 5\% of generated descriptions match ICD codes exactly; with 1, 3, and 5 few shots, this number rises to 32\%, 51\%, and 63\% respectively. For the non-instruct fine-tuned version of the model, the number of exact matches rises even further to 56\%, 70\% and 76\%.
This indicates an adaption to the task through the provided few-shot examples. 
Furthermore, it shows that the model has seen the respective ICD short or long descriptions during pre-training and can correctly replicate them when it is provided with enough context.
Among 5-shot predictions of Qwen2-72B Instruct, directly matched codes are correct 41\% of the time, while mapped ones reach 20\%. 
For Qwen2-7B, mapped codes retain 94\% of the accuracy of directly matched ones (30\% vs. 32\%). 
While the accuracy of mapped codes is lower, their generated text contains enough relevant information that enables the class-mapper to resolve the correct code.
This gap in performance is closely tied to the quality of the generated text. Accurate mappings result from meaningful clinical descriptions (e.g., \textit{Hypertensive heart disease}, \textit{Hypotension}), while nonsensical outputs (e.g., \textit{diagnosis 1}, \textit{diagnosis 2}, \textit{\_\_\_} etc.) almost never yield correct codes.


\begin{table}[htb!]
\centering
\resizebox{\linewidth}{!}{%
\begin{tabular}{lrr} 
\toprule
\textbf{Model}        & \textbf{\#Parsable} & \textbf{Avg. \#Diag.}  \\ 
\midrule
Qwen2 7B               & 0.00\%          & -                   \\ 
Llama 3.1 8B           & 0.00\%          & -                   \\ 
Mistral 7B             & 0.00\%          & -                   \\ 
Mistral Nemo           & 0.00\%          & -                   \\ 
Qwen2 72B              & 0.00\%          & -                   \\ 
Llama 3.1 70B          & 0.00\%          & -                   \\ 
\midrule
Qwen2 7B Instruct      & 97.25\%        & 19.91             \\ 
Llama 3.1 8B Instruct  &  0.49\%        & -                 \\ 
Mistral 7B Instruct    &  99.76\%       & 18.16             \\
Mistral Nemo Instruct  &  0.07\%        & 20.00             \\
Qwen2 72B Instruct     & 100.00\%       & 20.12             \\ 
Llama 3.1 70B Instruct &  3.82\%        & 20.16             \\ 
\end{tabular}
}
\label{table:json-performance}
\caption{Capability of evaluated models to produce valid JSON for the ICD-10 HOSP dataset in the zeroshot scenario (5493 examples). Non-instruct fine-tuned models all fail to generate valid JSON. Model size only has a small effect on the validity of the generated output in comparison to instruct-finetuning (Llama 3.1 vs Qwen2 or Mistral 7B vs Mistral Nemo).}
\end{table}

\begin{table}[htb]
\centering
\resizebox{\linewidth}{!}{%
\begin{tabular}{llll} 
\toprule
\textbf{Model}              & \textbf{Recall} & \textbf{Precision} & \textbf{MAP}  \\ 
\midrule
all-mpnet-base-v2            & 9.99            & 6.02               & 10.94         \\ 
S-PubMedBERT MSMarco & 10.61           & 6.31               & 10.71         \\ 
NeuML PubMedBERT            & \textbf{10.72}           & \textbf{6.92}               & \textbf{11.82}         \\
\bottomrule
\end{tabular}
}
\label{table:class-mapper-performance}
\caption{Average diagnosis prediction performance over all datasets and all experiments using different class mapping models in macro averaged metrics. NeuML PubMedBERT works best. However, performance differences between the different class mapping models are rather small.}
\end{table}


\subsection{Performance drop for longer inputs.}
\begin{figure}[h!]
    \centering
    \includegraphics[width=0.48\textwidth]{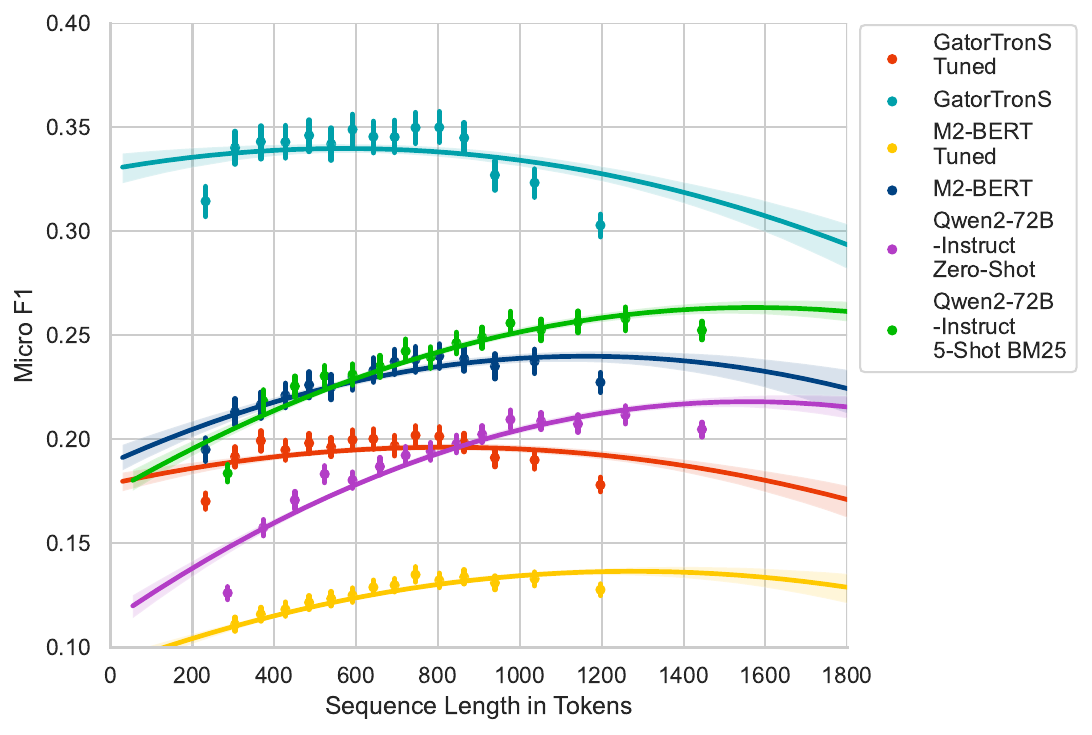}
    \caption{Micro $F_1$ over different sequence lengths averaged over all dataset splits. Encoder models have a reduced performance with sequences >850 tokens while the generative models tend have a more consistent performance on longer sequences. The token count varies across models due to differences in tokenizers.}
    \label{fig:token_count_reg}
\end{figure}

In \cref{fig:token_count_reg}, we demonstrate that longer sequence inputs correlate with a decrease in micro $F_1$ performance.
As anticipated, this decline is more pronounced for the encoder-based models BioMedBERT and GatorTronS, which are limited to sequence lengths of 512 tokens.
Models that are capable of processing extended sequences like the chosen generative models and the M2 encoder, exhibit only slight decreases in performance on longer sequences.
This highlights the necessity for long-context-capable domain-specific models.

\newpage

\subsection{ICU Results}

\begin{table}[h]
\resizebox{\linewidth}{!}{%
\begin{tabular}{clrrrrr} 
\toprule
\textbf{Data}                & \textbf{Retriever}                          & \textbf{\#FS} & \textbf{Rec.} & \textbf{Prec.} & \textbf{MAP}   & \textbf{MD Acc.}  \\ 
\midrule
\multirow{16}{*}{\rotatebox{90}{ICU}}
    & Zero-Shot prompted & 0 & \textbf{24.11} & 14.20 & 16.82 & 48.11  \\
    \cline{2-7}
    & \multirow{3}{*}{\textit{Random}} & 1 & 17.67 & 13.95 & 16.06 & 40.80  \\
    &                         & 3 & 15.90 & 13.27 & 16.00 & 41.22  \\
    &                         & 5 & 14.24 & 13.45 & 15.77 & 40.66  \\
\cline{2-7}
    & \multirow{3}{*}{BM25} & 1 & \uline{19.97} & 15.29 & 17.46 & 47.27  \\
    &                       & 3 & 18.89 & \textbf{15.89} & 19.09 & 51.18  \\
    &                       & 5 & 17.42 & \uline{15.73} & \textbf{20.48} & 52.06  \\
\cline{2-7}
     & \multirow{3}{*}{PubMedBERT$_{NeuML}$} & 1 & 18.25 & 13.83 & 17.16 & 45.59  \\
     &                                   & 3 & 17.13 & 13.97 & 18.43 & 50.50  \\
     &                                   & 5 & 16.10 & 14.48 & \uline{19.92} & \uline{52.34}  \\
\cline{2-7}
    & \multirow{3}{*}{BioMedBERT$_{ft}$}        & 1 & 18.97 & 14.60 & 17.69 & 44.80  \\
    &                                   & 3 & 18.04 & 14.86 & 18.96 & 50.92  \\
    &                                   & 5 & 18.12 & 15.44 & 19.57 & \textbf{52.59}  \\ 
\cline{2-7}
    & \multirow{3}{*}{\textit{Gold Heuristic}}             & 1 & 21.31 & 17.79 & 19.75 & 48.45  \\
    &                                   & 3 & 22.72 & 20.24 & 26.88 & 59.15  \\
    &                                   & 5 & 22.83 & 21.21 & 29.33 & 61.66  \\
\bottomrule
\end{tabular}
}
\caption{Few-shot performance of Qwen2-72B-Instruct averaged over ICD-9 and ICD-10 ICU splits using the depicted retrieval models. Metrics are macro averaged. BioMedBERT$_{ft}$ is a custom fine-tuned encoder model on the respective training split of the dataset. (\textbf{\#FS}: number of few-shot demonstrations, \textbf{MD Acc.}: Main diagnosis accuracy)}
\label{table:few_shot_results_icu}
\end{table}

\begin{figure}[h]
    \centering
    \includegraphics[width=0.48\textwidth]{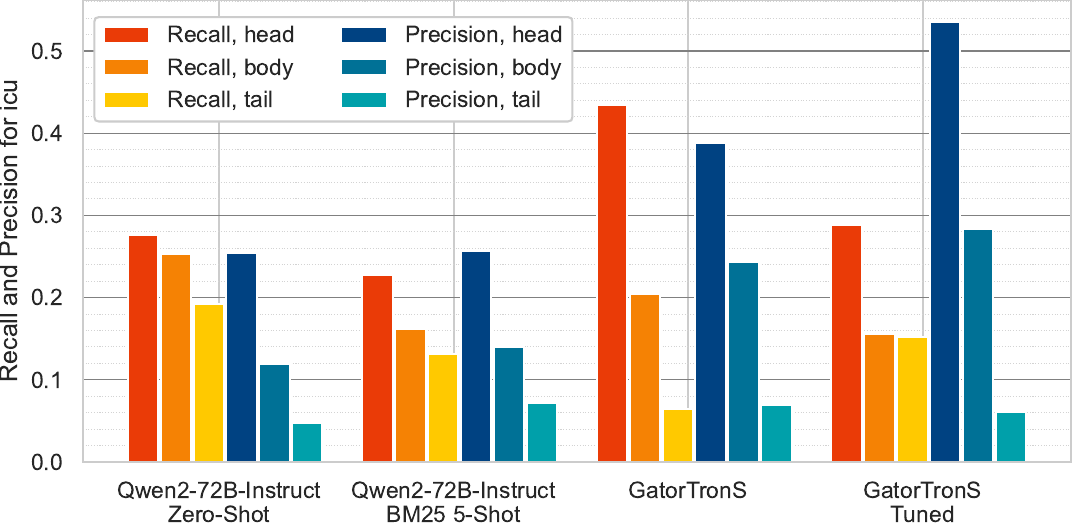}
    \caption{Macro recall and precision by model and classes grouped into tertiles by class frequency aggregated over both ICU splits (ICD-9 and ICD-10).}
    \label{fig:bar_class_frequency_icu}
\end{figure}

\begin{table*}
	[h!]
	\centering
	\resizebox{\linewidth}{!}{
		\begin{tabular}{clrrrrr|rrrrr|rrrrr}
			\toprule                                       &                              & \multicolumn{5}{c|}{\textbf{Zero-Shot prompted}} & \multicolumn{5}{c|}{\textbf{Zero-Shot prompted + CoT}} & \multicolumn{5}{c}{\textbf{Few-Shot 5 Demonstrations}} \\
			\multicolumn{1}{l}{\textbf{Data}} & \textbf{Model}            & \textbf{Rec.}     & \textbf{Prec.}    & \textbf{MAP}      & \textbf{MD Acc.}  & \textbf{F1} & \textbf{Rec.}     & \textbf{Prec.}    & \textbf{MAP}      & \textbf{MD Acc.}  & \textbf{F1}           & \textbf{Rec.}     & \textbf{Prec.}    & \textbf{MAP}      & \textbf{MD Acc.} & \textbf{F1} \\
			\midrule \multirow{24}{*}{\rotatebox{90}{ICU}} & \textbf{Encoder-based Models} \\
			& BioMedBERT-110M              & 19.60                                   & 17.99                                         & 32.44                                                 & 72.83             & 22.00            & \multicolumn{5}{c|}{--} & \multicolumn{5}{c}{--} \\
			& BioMedBERT-110M-tuned        & 15.56                                   & 23.13                                         & 20.03                                                 & 55.91             &  20.71            & \multicolumn{5}{c|}{--} & \multicolumn{5}{c}{--} \\
			& GatorTronS-345M              & \textbf{28.61}                          & 23.96                                         & \textbf{35.66}                                        & \textbf{78.44}    &  \textbf{25.77}           & \multicolumn{5}{c|}{--} & \multicolumn{5}{c}{--} \\
			& GatorTronS-345M-tuned        & 20.00                                   & \textbf{29.62}                                & 21.13                                                 & 63.66             &  25.34           & \multicolumn{5}{c|}{--} & \multicolumn{5}{c}{--} \\
			& M2-80M                       & 10.16                                   & 08.27                                         & 23.12                                                 & 56.05             &  9.50           & \multicolumn{5}{c|}{--} & \multicolumn{5}{c}{--} \\
			& M2-80M-tuned                 & 08.13                                   & 08.98                                         & 10.72                                                 & 35.05             &  8.34           & \multicolumn{5}{c|}{--} & \multicolumn{5}{c}{--} \\
			\cdashline{2-17}                               & \textbf{Generative LLM}       \\
			                                  & Llama 3.1-8B              & 13.97             & 14.81             & 13.50             & 33.43             & 11.20       & 12.25             & 13.41             & 12.36             & 35.57             & 09.76                 & 14.12             & 14.60             & 17.62             & 48.33         & 13.20 \\
			                                  & Llama 3.1-8B-Instruct     & 19.59             & 13.93             & 15.63             & 47.34             & 13.02       & 17.88             & 13.00             & 13.21             & 47.65             & 11.49                 & 15.52             & 14.36             & 15.93             & 44.27         & 13.32 \\
			                                  & Llama 3.1-70B             & 15.40             & 12.44             & 12.30             & 33.60             & 10.48       & 18.73             & 11.87             & 12.98             & 40.81             & 11.22                 & 13.74             & 12.10             & 16.54             & 50.85         & 12.13 \\
			                                  & Llama 3.1-70B-Instruct    & 20.31             & 13.36             & 14.70             & 50.37             & 12.55       & 21.24             & 12.66             & 15.85             & \underline{50.18} & 12.08                 & 17.06             & 15.38             & 16.24             & 44.90         & 14.47 \\
			                                  & Mistral-7B                & 08.84             & 13.49             & 09.94             & 25.66             & 08.08       & 00.34             & 02.63             & 04.58             & 01.38             & 00.12                 & 11.08             & 11.92             & 14.40             & 43.16         & 10.79 \\
			                                  & Mistral-7B-Instruct       & 23.65             & 13.67             & \underline{16.86} & \underline{51.37} & 13.69       & 21.59             & 12.12             & 14.64             & 46.10             & 12.30                 & \underline{20.96} & 15.08             & 16.98             & 47.72         & \underline{14.96} \\
			                                  & Mistral-Nemo-13B          & 13.49             & \underline{14.31} & 13.55             & 31.63             & 10.86       & 11.11             & 11.63             & 10.86             & 37.51             & 08.23                 & 11.31             & 12.73             & 15.13             & 44.95         & 11.00 \\
			                                  & Mistral-Nemo-13B-Instruct & 19.64             & 11.97             & 13.87             & 48.39             & 11.58       & 19.03             & 11.96             & 12.46             & 48.54             & 10.78                 & 11.24             & 13.35             & 15.24             & 43.28         & 11.37 \\
			                                  & Qwen2-7B                  & 20.95             & 13.92             & 15.54             & 40.97             & 13.11       & 19.05             & 13.96             & 15.49             & 40.19             & 12.55                 & 11.23             & 13.36             & 15.80             & 43.35         & 11.35 \\
			                                  & Qwen2-7B-Instruct         & 23.06             & 14.14             & 15.33             & 45.86             & 14.10       & 22.13             & 13.89             & 15.18             & 46.30             & 13.64                 & 18.23             & 14.40             & 17.61             & 50.52         & 14.15 \\
			                                  & Qwen2-72B                 & 22.55             & 14.06             & 15.89             & 43.28             & 13.70       & 20.32             & 13.14             & 15.57             & 40.93             & 12.56                 & 15.27             & 13.80             & 17.39             & 49.40         & 13.31 \\
			                                  & Qwen2-72B-Instruct        & \underline{24.11} & 14.20             & 16.82             & 48.11             & \underline{14.41}       & \underline{22.78} & \underline{14.10} & \underline{16.47} & 47.16             & \underline{14.12}                 & 17.42             & \underline{15.74} & \underline{18.75} & \uline{52.06} & 14.77 \\
			\bottomrule
		\end{tabular}
	}
	\caption{(\textbf{MAP}: \textit{Mean Average Precision}, \textbf{MD Acc.}: \textit{Main
		Diagnosis Accuracy}) Zero-shot performance on the ICU split over the top-20 predictions.
		The reported metrics are macro averaged. We aggregate ICD-9 and ICD-10 and report
		an unweighted mean. The dashed lines separate the encoder from the generative
		LLMs. CoT is not beneficial while instruct fine-tuning boosts performance.
		Threshold tuning decreases the performance of encoder models within the top-20
		predictions. We depict the best performing models in bold and underline the best
	performing generative models in each category.}
	\label{table:big_results_table_icu}
\end{table*}


\clearpage
\newpage
\begin{figure*}[ht]
    \centering
    \includegraphics[width=\linewidth]{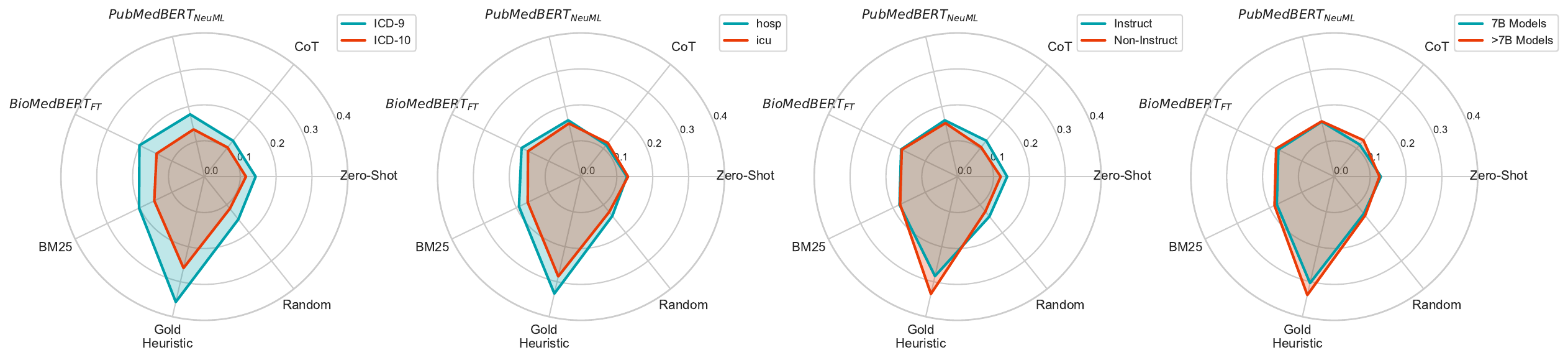}
    \caption{MAP for all generative models (Zero-Shot, CoT, and 5-shot with the respective retriever) aggregated for following criteria: Type of ICD codes, tasks, instruction and model size.}
    \label{fig:rad_map}
\end{figure*}

\begin{figure*}[ht]
    \centering
    \includegraphics[width=\linewidth]{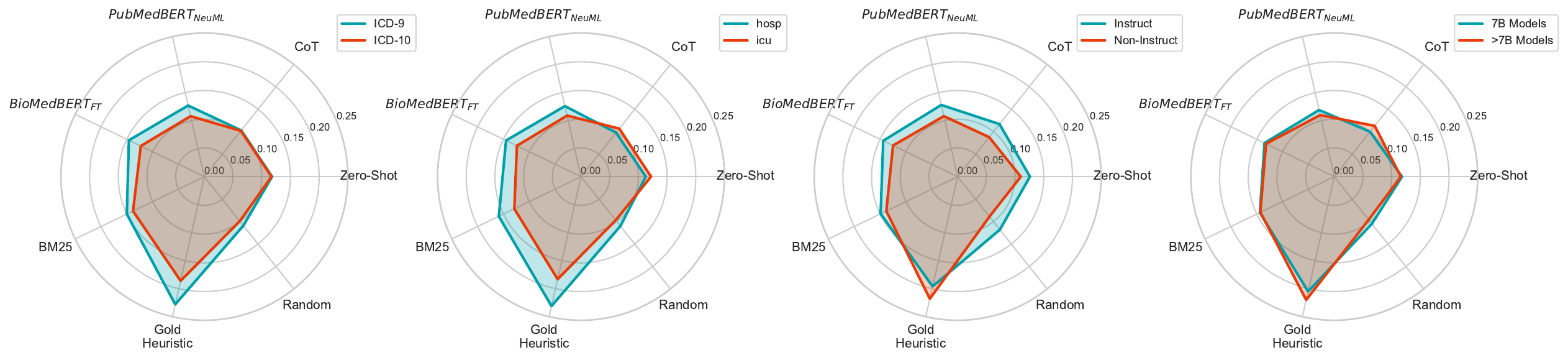}
    \caption{Macro F1 for all generative models (Zero-Shot, CoT, and 5-shot with the respective retriever) aggregated for following criteria: Type of ICD codes, tasks, instruction and model size.}
    \label{fig:rad_f1}
\end{figure*}

\begin{figure*}[h]
    \centering
    \includegraphics[width=\linewidth]{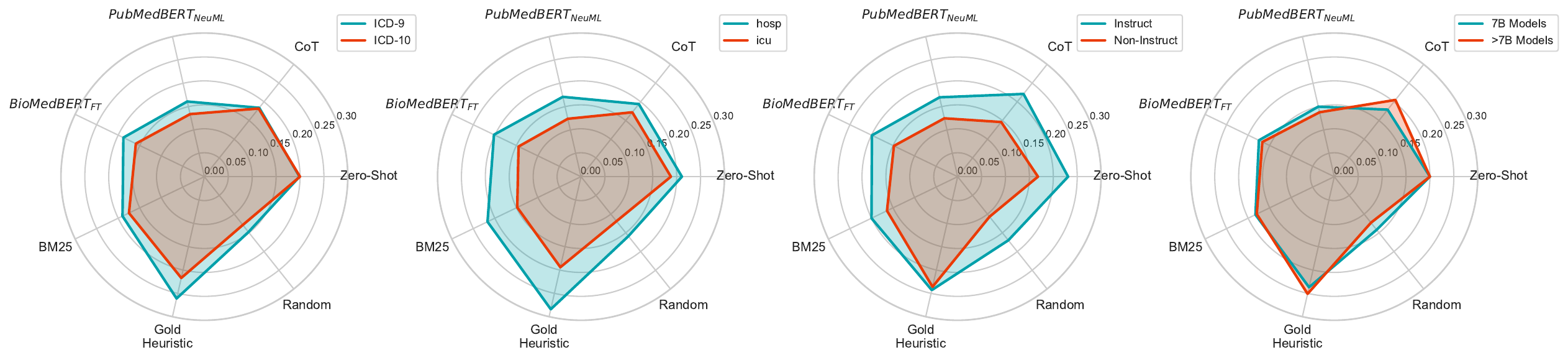}
    \caption{Macro recall for all generative models (Zero-Shot, CoT, and 5-shot with the respective retriever) aggregated for following criteria: Type of ICD codes, tasks, instruction and model size.}
    \label{fig:rad_recall}
\end{figure*}

\begin{figure*}[h]
    \centering
    \includegraphics[width=\linewidth]{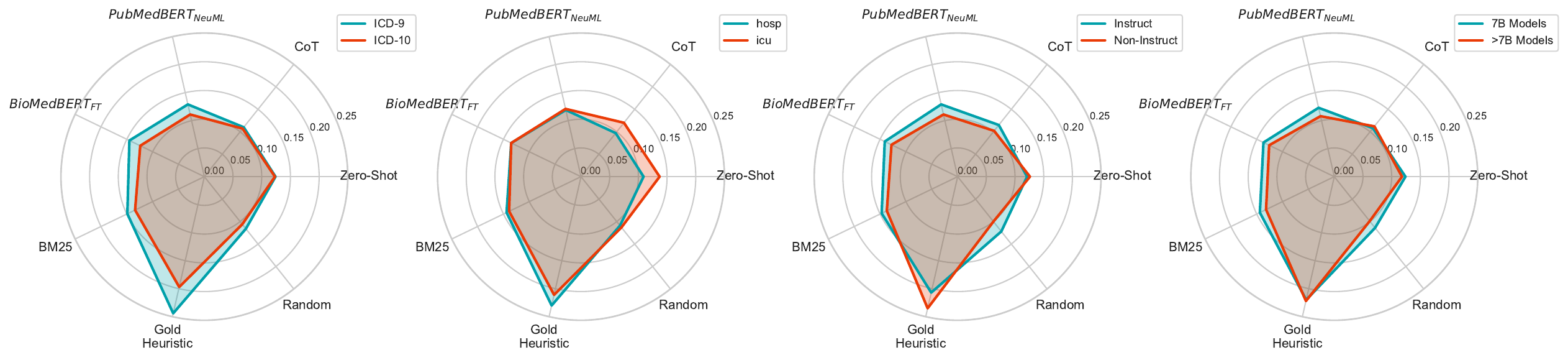}
    \caption{Macro precision for all generative models (Zero-Shot, CoT, and 5-shot with the respective retriever) aggregated for following criteria: Type of ICD codes, tasks, instruction and model size.}
    \label{fig:rad_precision}
\end{figure*}

\end{document}